\documentclass[letterpaper]{article} 
\usepackage{aaai2026}  
\usepackage{times}  
\usepackage{helvet}  
\usepackage{courier}  
\usepackage[hyphens]{url}  
\usepackage{graphicx} 
\usepackage{natbib}  
\usepackage{caption} 
\usepackage{amsmath}
\title{Schema Matching on Graph: Iterative Graph Exploration for Efficient and Explainable Data Integration}

\author{
    Mingyu Jeon,
    Jaeyoung Suh,
    Suwan Cho
}
\affiliations{
    MODULABS\\
    jkmcoma7@gmail.com, tjwodud04@gmail.com, swannyy7@gmail.com
}

\begin{document}
\maketitle

\begin{abstract}
Schema matching is a critical task in data integration, particularly in the medical domain where disparate Electronic Health Record (EHR) systems must be aligned to standard models like OMOP CDM. While Large Language Models (LLMs) have shown promise in schema matching, they suffer from hallucination and lack of up-to-date domain knowledge. Knowledge Graphs (KGs) offer a solution by providing structured, verifiable knowledge. However, existing KG-augmented LLM approaches often rely on inefficient complex multi-hop queries or storage-intensive vector-based retrieval methods.
This paper introduces SMoG (Schema Matching on Graph), a novel framework that leverages iterative execution of simple 1-hop SPARQL queries, inspired by successful strategies in Knowledge Graph Question Answering (KGQA). SMoG enhances explainability and reliability by generating human-verifiable query paths while significantly reducing storage requirements by directly querying SPARQL endpoints. Experimental results on real-world medical datasets demonstrate that SMoG achieves performance comparable to state-of-the-art baselines, validating its effectiveness and efficiency in KG-augmented schema matching.
\end{abstract}

\section{Introduction}

Schema matching, the task of identifying correspondence between schemas of different data sources, is a fundamental problem in data integration \cite{rahm2001survey}. In the medical domain, transforming disparate Electronic Health Record (EHR) systems into a standard data model, such as the OMOP Common Data Model (CDM), is a prerequisite for multi-center clinical research \cite{hripcsak2015observational, overhage2012validation}. However, real-world medical systems often contain hundreds of attributes with opaque names, synonyms, and acronyms, making manual matching impractical and error-prone \cite{kahn2016harmonized}.

To address this, various schema matching approaches have been proposed. Traditional methods are categorized into Schema-based \cite{rahm2001survey, do2002coma}, which utilize schema metadata (e.g., attribute names, types), and Instance-based \cite{kang2003schema}, which analyze the distribution or patterns of actual data values. Schema-based methods like Cupid \cite{madhavan2001generic} and COMA \cite{do2002coma} combine linguistic and structural similarities but suffer performance degradation when attribute names are opaque \cite{doan2001reconciling}. Instance-based methods analyze statistical distributions \cite{larson1989theory} or query logs \cite{elmeleegy2008usage}, but their application in the medical domain is often limited due to restricted data access and privacy concerns \cite{vatsalan2013privacy}.

Recently, machine learning-based approaches utilizing Pre-trained Language Models (PLMs) have emerged \cite{devlin2019bert, liu2019roberta}. Models like SMAT \cite{zhang2021smat} and Unicorn \cite{dong2023unicorn} fine-tune BERT or DeBERTa to classify attribute pairs. However, these methods require large-scale labeled data and lack domain-specific knowledge to capture complex medical terminology relationships \cite{peeters2023supervised}. The advent of Large Language Models (LLMs) has opened possibilities for zero-shot or few-shot schema matching \cite{brown2020language, ouyang2022training}. While models like Jellyfish \cite{narayan2024can} and Prompt-Matcher \cite{anonymous2024prompt} have demonstrated the potential of LLMs, they still face limitations such as hallucination \cite{ji2023survey} and a lack of up-to-date domain knowledge \cite{huang2023survey}.

To mitigate these limitations, research leveraging Knowledge Graphs (KGs) \cite{pan2024unifying, yasunaga2021qa}, which provide explicit and editable structured knowledge, is gaining attention. Medical-specific KGs like SNOMED-CT \cite{donnelly2006snomed} and UMLS \cite{bodenreider2004unified} are particularly effective in augmenting LLM reasoning. A state-of-the-art study in this field, KG-RAG4SM \cite{ma2025knowledgegraphbasedretrievalaugmentedgeneration}, applied Retrieval-Augmented Generation (RAG) to schema matching for the first time. It proposed various retrieval methods, including Vector-based and Query-based approaches, achieving significant F1-Score improvements on benchmark datasets like MIMIC \cite{johnson2016mimic} and CMS \cite{cms2020data}.

However, KG-RAG4SM \cite{ma2025knowledgegraphbasedretrievalaugmentedgeneration} exhibited a critical limitation. The authors concluded that the Query-based subgraph retrieval method "does not work well in practice due to the high computational cost and time required in large-scale KGs" \cite{ma2025knowledgegraphbasedretrievalaugmentedgeneration}. Citing poor quality of LLM-generated queries, inefficiency of complex multi-hop queries, and incomplete retrieval, they dismissed the Query-based approach and adopted Vector-based triple retrieval. Yet, Vector-based retrieval relies solely on semantic similarity, possessing a fundamental limitation in \textbf{explicitly distinguishing or traversing specific paths} based on structural relationships like \texttt{subclass\_of} or \texttt{different\_from}.

\textbf{In addition to these limitations of the Vector-based approach adopted by KG-RAG4SM \cite{ma2025knowledgegraphbasedretrievalaugmentedgeneration}, this study challenges their conclusion dismissing the 'Query-based' approach.} We argue that the problem lies not in the Query-based approach itself, but in the design of 'complex multi-hop SPARQL queries'. Indeed, ToG (Think on Graph) \cite{sun2023think}, a successful study in Knowledge Graph Question Answering (KGQA), successfully explores KGs by iteratively executing 'template-based simple 1-hop SPARQL queries'.

Based on this insight, we propose the \textbf{SMoG (Schema Matching on Graph)} framework. Instead of following the existing Vector-based RAG approach, SMoG performs schema matching through the iterative execution of 1-hop SPARQL queries, inspired by ToG \cite{sun2023think}. This approach offers two distinct advantages. First, the knowledge retrieval process is explicit, transparent, and human-verifiable, thereby enhancing the reliability and explainability of the results. Second, it directly accesses SPARQL endpoints without the need to embed the entire KG or store massive vector indices, making it highly efficient in terms of knowledge base management and storage space.

The main contributions of this study are as follows:
\begin{itemize}
    \item \textbf{Re-establishing Query-based Schema Matching:} We are the first to identify that the failure of the 'Query-based' knowledge retrieval method, deemed inefficient by existing SOTA research (KG-RAG4SM \cite{ma2025knowledgegraphbasedretrievalaugmentedgeneration}), stems from the 'complex multi-hop query' design rather than the approach itself.
    \item \textbf{Proposal of SMoG Framework:} Inspired by ToG \cite{sun2023think} in KGQA, we propose the \textbf{SMoG framework}, which adapts \textbf{'iterative execution of template-based 1-hop SPARQL queries'} to the Schema Matching (SM) task. This is a novel approach in schema matching that performs matching through SPARQL query exploration.
    \item \textbf{Ensuring Explainability and Reliability:} Instead of relying on the 'black-box' nature of Vector-based retrieval or LLM 'hallucinations', SMoG explicitly generates human-verifiable 1-hop query exploration paths. This increases result reliability and facilitates debugging.
    \item \textbf{Proving Storage Efficiency and Practicality:} Unlike the Vector-based RAG \cite{ma2025knowledgegraphbasedretrievalaugmentedgeneration} method that requires embedding the entire KG and maintaining massive vector indices, SMoG queries SPARQL endpoints directly. We demonstrate that it achieves performance comparable to SOTA baselines on real-world medical datasets (CMS) while securing \textbf{superior storage efficiency}.
\end{itemize}

The remainder of this paper is organized as follows. Section 2 reviews related work, and Section 3 details the query strategy and iterative execution mechanism of the proposed SMoG framework. Section 4 presents the experimental setup, baselines, ablation study, and result analysis, followed by the conclusion and future research directions in Section 5.

\section{Related Work}

\subsection{Schema Matching}

\subsubsection{Traditional Schema Matching Methods}
Traditional schema matching research is broadly divided into Schema-based and Instance-based methods. Schema-based methods utilize schema metadata such as attribute names, data types, and constraints \cite{rahm2001survey, do2002coma}. Early studies in this field systematically classified and surveyed schema matching methodologies \cite{rahm2001survey}. Representative works include Cupid \cite{madhavan2001generic}, which combined linguistic similarity (WordNet-based) and structural similarity (tree matching), and COMA \cite{do2002coma}, which improved accuracy by integrating results from multiple matchers through meta-matching. Similarity Flooding \cite{melnik2002similarity} performed matching through graph-based fixed-point computation.

On the other hand, Instance-based methods analyze the distribution and patterns of actual data values \cite{kang2003schema}. iMAP \cite{dhamankar2004imap} estimated schema matching probabilities via Bayesian learning. However, these methods are difficult to apply in domains where instance-level access is restricted due to privacy constraints, such as medical data \cite{vatsalan2013privacy}.

Additionally, Usage-based methods utilize query logs to analyze relationships between attributes \cite{elmeleegy2008usage}. Elmeleegy et al. (2008) extracted patterns of co-occurrence, joins, and aggregate function usage between attributes from query logs \cite{elmeleegy2008usage}. However, this approach has a clear limitation in that it is inapplicable in environments with new systems or insufficient query logs.

\subsubsection{Machine Learning-based Schema Matching}
Recently, machine learning-based methods utilizing Pre-trained Language Models (PLMs) have gained attention \cite{devlin2019bert, liu2019roberta}. SMAT \cite{zhang2021smat} learned the similarity of attribute pairs using Siamese Networks and GloVe/BERT embeddings. Unicorn \cite{dong2023unicorn} fine-tuned the DeBERTa model to perform schema matching as an attribute pair classification problem. However, these PLM-based methods require large-scale labeled training data and have limitations in capturing the subtle semantic relationships of complex domain-specific terms (e.g., medical terminology) \cite{peeters2023supervised}.

\subsubsection{LLM-based Schema Matching}
The emergence of Large Language Models (LLMs) has opened new possibilities in the field of schema matching. LLMs can perform schema matching via zero-shot or few-shot learning based on vast pre-trained knowledge \cite{brown2020language, ouyang2022training}. Jellyfish \cite{narayan2024can} achieved high accuracy by fine-tuning StarCoder, a Code LLM, for schema matching tasks. ReMatch \cite{anonymous2024rematch} utilized Retrieval-Augmented Generation (RAG) to retrieve similar matching examples and construct few-shot prompts. Matchmaker \cite{anonymous2024matchmaker} iteratively improved matching performance through self-improving LLM programs, and Prompt-Matcher \cite{anonymous2024prompt} proposed a systematic prompting strategy utilizing JSON schema structures.

However, these LLM-based approaches are not free from the inherent limitation of LLMs, the hallucination problem \cite{ji2023survey}, which can lead to the generation of non-existent relationships or incorrect matching due to a lack of up-to-date domain knowledge \cite{huang2023survey}.

\subsection{Knowledge Graph-augmented LLM}

\subsubsection{Background of KG-augmented LLM}
To mitigate the aforementioned limitations of LLMs (hallucination, lack of recency, opaque decision-making), utilizing Knowledge Graphs (KGs) is being actively researched \cite{ji2023survey, huang2023survey}. KGs offer the advantage of providing explicit and editable structured knowledge, with easy updates for the latest information \cite{pan2024unifying, yasunaga2021qa}. KG-augmented LLMs aim to improve the accuracy and reliability of LLMs by integrating this verified knowledge from KGs into the LLM's reasoning process \cite{baek2023knowledge, sun2023think}.

\subsubsection{KG-augmented LLM for Reasoning}
Research on utilizing KGs for LLM reasoning is prominent in the Knowledge Graph Question Answering (KGQA) field. Among them, \textbf{Think-on-Graph (ToG) \cite{sun2023think}} is a pioneering study that inspired the core idea of this research. ToG proposed a method where, instead of generating complex multi-hop queries at once, the LLM iteratively executes \textbf{'template-based simple 1-hop SPARQL queries'} to explore the KG. At each step (hop), the LLM decides the next exploration action based on the retrieved triplets and searches for promising reasoning paths via beam search. This approach lowers the risk of complex query generation failure and makes the exploration process transparent.

Following ToG, Reasoning on Graphs (RoG) \cite{luo2023reasoning} conducted research on performing more faithful and interpretable reasoning based on KG paths. Furthermore, \textbf{Plan-on-Graph (PoG) \cite{chen2024plan}} extended ToG's idea by introducing self-correcting adaptive planning, where the LLM dynamically modifies the plan, achieving SOTA performance in KGQA tasks.

\subsubsection{KG-augmented LLM for Schema Matching}
The SOTA research utilizing KGs in the schema matching field is \textbf{KG-RAG4SM \cite{ma2025knowledgegraphbasedretrievalaugmentedgeneration}}. This study was the first attempt to apply RAG to schema matching, proposing various subgraph retrieval methods such as Vector-based, Query-based, and BFS-based. It demonstrated the validity of utilizing KGs by achieving F1-Score improvements over existing LLM-based methods on medical datasets like MIMIC, Synthea, and CMS.

However, the researchers of KG-RAG4SM \cite{ma2025knowledgegraphbasedretrievalaugmentedgeneration} pointed out a clear limitation regarding the 'Query-based' retrieval method. They concluded that "Cypher queries generated by LLMs on large-scale KGs are computationally expensive and time-consuming, making them inefficient" \cite{ma2025knowledgegraphbasedretrievalaugmentedgeneration}. Citing the degradation of LLM-generated query quality and the inefficiency of multi-hop queries, they ultimately adopted Vector-based triple retrieval as their final model. However, this method bears the burden of massive storage space and computational costs required to embed the entire KG and build/maintain a vast vector index. Moreover, this approach has a fundamental limitation in that it is difficult to explain the matching results or trust the path due to the 'black-box' nature of Vector-based retrieval.

\subsection{Position of This Research}
Existing studies introduced KG-RAG to solve the hallucination problem of LLMs in schema matching \cite{ma2025knowledgegraphbasedretrievalaugmentedgeneration}, but concluded that the Query-based approach is inefficient and relied on Vector-based RAG.

This study directly challenges this conclusion. We argue that the cause of failure in KG-RAG4SM \cite{ma2025knowledgegraphbasedretrievalaugmentedgeneration} lies not in the 'Query-based' approach itself, but in the design of 'complex multi-hop queries'.

Therefore, we propose the \textbf{SMoG (Schema Matching on Graph)} framework, which adapts the strategy of \textbf{'iterative execution of simple 1-hop queries'}, proven in ToG \cite{sun2023think} of the KGQA field, to the schema matching task. Unlike Vector-based RAG, SMoG does not require KG embedding and explores explainable and reliable matching paths by borrowing ToG's approach.

\section{Methodology}

\begin{figure*}[t]
    \centering
    \includegraphics[width=0.95\textwidth]{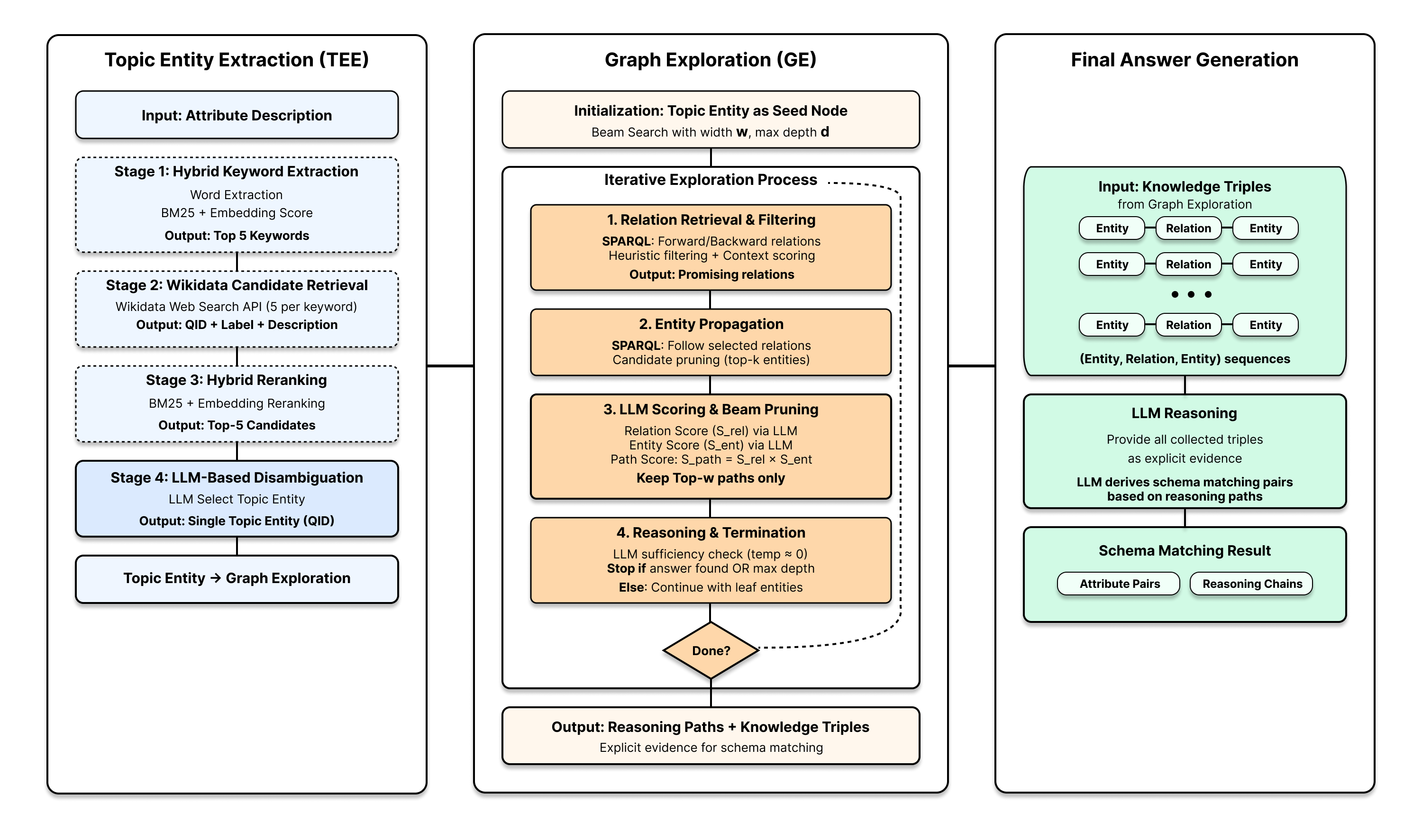}
    \caption{Overview of the SMoG Framework. The framework consists of two main phases: Topic Entity Extraction (TEE) and Graph Exploration (GE). TEE identifies the starting entity for a given attribute, while GE iteratively explores the Knowledge Graph using 1-hop SPARQL queries to find the optimal matching path.}
    \label{fig:smog_architecture}
\end{figure*}

\subsection{SMoG Architecture Overview}
The SMoG (Schema Matching on Graph) framework is designed to overcome the limitations of existing Vector-based RAG approaches by providing an explainable and efficient schema matching process. As illustrated in Figure \ref{fig:smog_architecture}, the framework comprises two core processes: \textbf{Topic Entity Extraction (TEE)} and \textbf{Graph Exploration (GE)}.

First, the \textbf{Topic Entity Extraction (TEE)} module aims to accurately identify a \textbf{Topic Entity} within the Knowledge Graph (KG) that represents the semantics of the target attribute. Using the attribute's description as input, the module leverages LLM-based reasoning to extract the most descriptive Topic Entity candidate from the KG. These extracted entities serve as the starting points for the subsequent exploration phase.

Second, the \textbf{Graph Exploration (GE)} module collects structural and semantic knowledge (Knowledge Triples) from the KG to answer the schema matching question, starting from the identified Topic Entity. Inspired by the Think-on-Graph (ToG) approach \cite{sun2023think} in KGQA, SMoG employs an iterative execution of simple 1-hop SPARQL queries instead of generating complex multi-hop queries. The LLM infers the necessary relations for the next hop based on the currently retrieved triples, iteratively accumulating information required for the answer. The exploration terminates when sufficient information is gathered to answer the question or when a predefined maximum depth is reached.

Finally, in the \textbf{Final Answer Generation} phase, all Knowledge Triples collected during the GE phase are provided to the LLM. Based on this explicit evidence, the LLM derives the final schema matching pair.

\subsection{Topic Entity Extraction (TEE)}

\subsubsection{Objective and Strategy}
The primary goal of TEE is to normalize the natural language description of a schema attribute into a single \textbf{Topic Entity (QID)} in Wikidata, providing a precise starting point for Graph Exploration. Due to the diversity of domain concepts (e.g., synonyms, acronyms) and semi-structured descriptions, simple keyword matching is often insufficient. To address this, we employ a hybrid strategy that combines \textbf{lexical and semantic signals} to refine candidates, followed by an \textbf{LLM-based disambiguation} step to ensure robustness and accuracy.

\subsubsection{TEE Pipeline}
The TEE process follows a "low-cost candidate convergence $\rightarrow$ high-confidence single selection" principle, consisting of four stages:

\paragraph{Stage 1: Hybrid Keyword Generation}
We generate a small set (Top-K=5) of representative keywords that satisfy both retrieval efficiency and semantic coverage. The input description is preprocessed (lowercasing, tokenization, stopword filtering) to create a candidate token set. We calculate a \textbf{BM25 score} to capture lexical signals (relative frequency and compressibility) and an \textbf{Embedding score} (cosine similarity between the token and the full description) to capture semantic signals. The final Top-5 keywords are selected based on a weighted sum ($0.4 \times \text{BM25} + 0.6 \times \text{Embedding}$), prioritizing semantic relevance.

\paragraph{Stage 2: Wikidata Candidate Retrieval}
Using the keywords from Stage 1, we retrieve an initial pool of entity candidates from Wikidata. We query the `wbsearchentities` API for each keyword and collect `id (QID)`, `label`, and `description` for the top-5 results. Candidates are deduplicated by QID, and their textual representation is unified into a `full\_text` field combining the label and description.

\paragraph{Stage 3: Hybrid Reranking}
To sort the candidates by global relevance to the original attribute description, we perform hybrid reranking. We calculate a \textbf{BM25 reranking score} using the description as a query against the candidate corpus, and an \textbf{Embedding reranking score} based on cosine similarity. The final score is again a weighted sum ($0.4 \times \text{BM25} + 0.6 \times \text{Embedding}$). Only the top-5 candidates are passed to the next stage to balance quality and cost.

\paragraph{Stage 4: LLM-Based Disambiguation}
The final selection relies on an LLM to discern subtle semantic or common-sense differences among the top candidates. The LLM receives the structured information of the top-5 candidates and determines the single most appropriate QID.

\subsection{Graph Exploration via SMoG}

\subsubsection{Objective and Overview}
The Graph Exploration (GE) module aims to discover the optimal \textbf{Reasoning Chain} in the KG that leads to the answer for the schema matching query, starting from the Topic Entities identified in the TEE phase. We adopt a \textbf{Beam Search}-based multi-hop reasoning framework. To efficiently control the vast search space of the graph, we maintain only the most promising paths (Top-$w$) at each step while extending the depth ($d$).

\subsubsection{Iterative Exploration Process}
The exploration iterates through the following steps until the maximum depth ($D_{max}$) is reached or an answer is found.

\paragraph{1. Relation Retrieval and Filtering}
First, we retrieve all adjacent relations for the current entity ($e$) using SPARQL. We query both forward (where $e$ is the subject) and backward (where $e$ is the object) relations. To improve efficiency, we apply heuristic filtering to exclude irrelevant metadata relations (e.g., `instance of`, `URL`). The remaining relations are prioritized based on the semantic similarity between the query intent and the relation label.

\paragraph{2. Entity Propagation}
Next, we propagate to the next hop's candidate entities by traversing the selected promising relations ($p$). We execute SPARQL queries to find entities connected via these relations. To prevent combinatorial explosion, if a relation connects to an excessive number of entities, we prune the candidates, keeping only the top-$k$ entities based on connection strength or importance.

\paragraph{3. LLM-based Scoring and Beam Pruning}
This step is crucial for selecting promising paths using the LLM's reading comprehension and reasoning capabilities.
\begin{itemize}
    \item \textbf{Relation Scoring ($S_{rel}$):} The LLM evaluates which of the candidate relations are most likely to lead to the answer, given the query and current entity. It assigns a confidence score (0-1) to each relation.
    \item \textbf{Entity Scoring ($S_{ent}$):} The LLM evaluates the likelihood that the entities reached via the selected relations are the answer or close to it.
    \item \textbf{Beam Pruning:} The final path score is defined as $S_{path} = S_{rel} \times S_{ent}$. We sort all candidate paths by $S_{path}$ and keep only the top-$w$ paths (Beam Width), pruning the rest. This ensures the computational cost remains linear with respect to depth ($O(w \times d)$).
\end{itemize}

\paragraph{4. Reasoning and Termination}
In this control step, we verify if the accumulated knowledge paths are sufficient to solve the query.
\begin{itemize}
    \item \textbf{Chain Aggregation:} The top-$w$ reasoning chains are aggregated into a single context.
    \item \textbf{Sufficiency Verification:} The LLM assesses whether the answer can be derived from the current context. We use a low temperature setting to minimize hallucination.
    \item \textbf{Decision:} If a clear answer is identified (\textbf{Stop Condition Met}), the loop terminates, and the final answer and reasoning path are returned. If information is insufficient, the leaf entities of the current paths become the seeds for the next hop, and the process repeats.
    \item \textbf{Fail-safe:} If the maximum depth is reached without a definite answer, a "Half-stop" strategy is triggered to generate a best-effort answer based on the partial information collected.
\end{itemize}
\section{Experiment}

\subsection{Experiment Setup}

\subsubsection{Dataset Description}
We conducted our experiments on the \textbf{CMS (Centers for Medicare \& Medicaid Services)} dataset, a real-world medical dataset characterized by complex schema mappings and high domain specificity. To ensure fair comparison and reproducibility with state-of-the-art baselines, \textbf{we utilized the specific CMS Test dataset provided by the KG-RAG4SM \cite{ma2025knowledgegraphbasedretrievalaugmentedgeneration} repository}. The test set consists of \textbf{2,563} attribute pairs. A critical characteristic of this dataset is its \textbf{extreme class imbalance}, with only \textbf{25 positive matches (0.97\%)} and \textbf{2,538 negative matches (99.03\%)}.

\subsubsection{Model \& Knowledge Graph}
We evaluated our proposed \textbf{SMoG (Schema Matching on Graph)} framework, which integrates a Knowledge Graph-augmented LLM approach.
\begin{itemize}
    \item \textbf{Knowledge Graph:} We employed \textbf{Wikidata} as the external knowledge source, consistent with the setup in KG-RAG4SM, to provide broad coverage of medical and general domain entities.
    \item \textbf{Model:} We utilized \textbf{GPT-4o-mini} as the core reasoning engine, enhanced with retrieval-augmented generation (RAG) capabilities.
\end{itemize}

\subsubsection{Evaluation Metrics}
Given the class imbalance, we report \textbf{Precision}, \textbf{Recall}, and \textbf{F1-Score} as the primary metrics.

\subsection{Results}

\subsubsection{Baseline Comparison}
We compare our approach against state-of-the-art baselines referenced in recent literature (e.g., KG-RAG4SM \cite{ma2025knowledgegraphbasedretrievalaugmentedgeneration}, Jellyfish \cite{narayan2024can}, SMAT \cite{zhang2021smat}).

\begin{table}[h]
\centering
\resizebox{\columnwidth}{!}{
\begin{tabular}{l|ccc}
\hline
\textbf{Model} & \textbf{Precision} & \textbf{Recall} & \textbf{F1-Score} \\ \hline
\textbf{Ours (SMoG)} & \textbf{43.48\%} & \textbf{40.00\%} & \textbf{41.67\%} \\
KG-RAG4SM \cite{ma2025knowledgegraphbasedretrievalaugmentedgeneration} & 52.38\% & 44.00\% & 47.82\% \\
Unicorn (Fine-tuned) \cite{dong2023unicorn} & 59.99\% & 35.99\% & 44.99\% \\
SMAT \cite{zhang2021smat} & 31.57\% & 48.00\% & 38.09\% \\
Jellyfish \cite{narayan2024can} & 30.00\% & 36.00\% & 32.72\% \\ \hline
\end{tabular}
}
\caption{Performance comparison on the CMS dataset.}
\label{tab:baseline_comparison}
\end{table}

\subsubsection{Validation of Core Contributions}
Our experimental results strongly support the core contributions proposed in this study:
\begin{enumerate}
    \item \textbf{Re-establishing Query-based Approaches:} SMoG achieved an \textbf{F1-score of 41.67\%}, which is highly competitive with the SOTA Vector-based method (KG-RAG4SM, 47.82\%) and outperforms LLM-based baselines (Jellyfish, 32.72\%). This empirical evidence refutes the previous assumption that query-based methods are inherently inferior, demonstrating that \textbf{"iterative 1-hop queries"} are a viable and effective strategy.
    \item \textbf{Efficiency \& Explainability:} By achieving comparable performance \textit{without} relying on massive vector indexes, SMoG validates its \textbf{storage efficiency}. Furthermore, the explicit reasoning chains (analyzed in Sec 4.3) demonstrate the \textbf{explainability} of our framework, offering a transparent alternative to the "black-box" nature of vector retrieval.
\end{enumerate}

\subsection{Analytic Study: Impact of Reasoning Chains and Topic Relevance}
To understand the model's decision-making process, we analyzed the intermediate outputs of the SMoG pipeline: \textbf{Topic Entity Scores} and \textbf{Reasoning Chain Counts}.

\subsubsection{Impact of Reasoning Chains: Depth Analysis}
We analyzed the \textbf{Average Depth of Reasoning Chains} (final selected paths) generated by the model to understand its decision-making behavior.

\begin{table}[h]
\centering
\resizebox{\columnwidth}{!}{
\begin{tabular}{l|c|c}
\hline
\textbf{Group} & \textbf{Count} & \textbf{Avg. Depth} \\ \hline
\textbf{Match / Correct (TP)} & 10 & \textbf{2.50} \\
\textbf{Non-Match / Correct (TN)} & 2522 & \textbf{2.48} \\
\textbf{Match / Incorrect (FN)} & 15 & \textbf{2.80} \\
\textbf{Non-Match / Incorrect (FP)} & 16 & \textbf{1.62} \\ \hline
\end{tabular}
}
\caption{Average depth of reasoning chains by prediction group.}
\label{tab:depth_analysis}
\end{table}

\textbf{Interpretation:}
\begin{itemize}
    \item \textbf{TP (2.50):} \textbf{Consistent Reasoning.} Deep chains validate true matches.
    \item \textbf{TN (2.48):} \textbf{Thorough Verification.} Active disproval of connections.
    \item \textbf{FN (2.80):} \textbf{Over-Thinking.} Excessive complexity leads to confusion.
    \item \textbf{FP (1.62):} \textbf{Premature Commitment.} Hasty acceptance of weak connections.
\end{itemize}

\textbf{Conclusion:} The contrast between \textbf{FN (2.80)} and \textbf{FP (1.62)} provides a critical insight: \textbf{Errors stem from opposite behaviors.} False Negatives arise from \textit{excessive complexity} (Over-thinking), while False Positives arise from \textit{insufficient verification} (Premature Commitment). Future work should focus on enforcing deeper verification for potential matches to reduce FPs.

\subsubsection{Analysis of Topic Entity Extraction (TEE) Strategy}
We performed a deep dive into the TEE scores by grouping the data based on the \textbf{Model's Prediction Outcome} to understand the correlation between TEE signals and final performance.

\begin{table*}[t]
\centering
\begin{tabular}{l|c|ccc|ccc}
\hline
\textbf{Group} & \textbf{Count} & \textbf{OMOP BM25} & \textbf{OMOP Emb} & \textbf{OMOP Score} & \textbf{CMS BM25} & \textbf{CMS Emb} & \textbf{CMS Score} \\ \hline
\textbf{TP} & 10 & 3.50 & 0.39 & \textbf{0.68} & 1.43 & 0.43 & 0.81 \\
\textbf{FN} & 15 & \textbf{5.59} & 0.40 & \textbf{0.72} & 1.50 & 0.40 & 0.80 \\
\textbf{TN} & 2522 & \textbf{5.88} & 0.39 & \textbf{0.76} & 1.89 & 0.41 & 0.81 \\
\textbf{FP} & 16 & 4.55 & \textbf{0.31} & 0.68 & 1.68 & 0.37 & 0.79 \\ \hline
\end{tabular}
\caption{TEE score analysis by prediction group.}
\label{tab:tee_analysis}
\end{table*}

\textbf{Interpretation:}
\begin{itemize}
    \item \textbf{TP:} \textbf{Semantic Alignment.} The highest \textbf{CMS Emb Score (0.43)} indicates that correct matches are driven by strong semantic similarity between the source description and the entity.
    \item \textbf{FN:} \textbf{Lexical Distraction.} Despite high lexical scores (OMOP Score 0.72), the model predicted Negative, suggesting that high keyword overlap might act as a distractor or indicate ambiguous candidates.
    \item \textbf{TN:} \textbf{Robust Rejection.} The highest overall scores (OMOP Score 0.76) in this group confirm that the model successfully rejects candidates even when they have high lexical overlap, overcoming the "keyword trap."
    \item \textbf{FP:} \textbf{Semantic Hallucination.} The lowest \textbf{CMS Emb Score (0.37)} reveals that the model forced a match despite weak semantic evidence.
\end{itemize}

\textbf{Key Insight: Lexical Paradox vs. Semantic Reliability.}
\begin{itemize}
    \item \textbf{Lexical Paradox:} High lexical overlap (OMOP BM25) and Total Scores are inversely correlated with Positive predictions. This confirms that in the medical domain, standardized terms often act as \textbf{"Lexical Distractors,"} and the model correctly exercises caution (Negative prediction) in these high-overlap cases.
    \item \textbf{Semantic Reliability:} In contrast, the \textbf{CMS Embedding Score} shows a direct positive correlation with correct matching (TP=0.43 vs. FP=0.37). This suggests that \textbf{semantic similarity} is the reliable signal for distinguishing true matches from hallucinations.
\end{itemize}

\section{Discussion}

Our study revisited the potential of query-based approaches in schema matching, a direction previously dismissed by state-of-the-art research due to efficiency concerns. By analyzing the performance and behavior of SMoG, we provide new perspectives on the "Query vs. Vector" debate and the nature of LLM reasoning in this domain.

\subsection{Re-evaluating the "Query-based" Hypothesis}
The primary motivation of this work was to challenge the conclusion of KG-RAG4SM \cite{ma2025knowledgegraphbasedretrievalaugmentedgeneration} that query-based methods are inherently impractical for large-scale KGs. Our results with SMoG demonstrate that the issue lay not in the query-based approach itself, but in the complexity of the queries. By adopting an iterative, 1-hop SPARQL strategy inspired by ToG \cite{sun2023think}, SMoG achieved competitive F1-scores (41.67\%) comparable to vector-based baselines, effectively refuting the notion that query-based methods cannot scale or perform. This confirms our hypothesis that a "divide-and-conquer" exploration strategy is a viable, and perhaps superior, alternative to complex multi-hop query generation.

\subsection{The Dual Nature of Reasoning Errors}
Beyond feasibility, our depth analysis revealed distinct behavioral patterns in LLM reasoning. We observed a dichotomy in error modes: False Negatives were associated with deeper reasoning chains ("Over-thinking," avg. depth 2.80), while False Positives stemmed from shallow, hasty conclusions ("Premature Commitment," avg. depth 1.62). This suggests that while the iterative query mechanism allows for deep exploration, it requires a regulatory mechanism to balance depth—preventing "reasoning drift" in complex cases while enforcing rigorous verification in seemingly obvious ones.

\subsection{Lexical Noise vs. Semantic Signal}
Our TEE analysis highlighted a critical distinction between lexical and semantic signals. We observed a \textbf{"Lexical Paradox"} where high lexical overlap (OMOP Score) often served as a distractor, correlating more with Non-Matches. The model's ability to reject these high-score candidates (TN group) demonstrates its robustness against lexical noise. Conversely, the \textbf{"Semantic Signal"} (CMS Embedding Score) proved to be the decisive factor for correct matching, with True Positives showing significantly higher semantic similarity than False Positives. This suggests that future schema matching research should prioritize \textbf{semantic alignment} over simple keyword matching and focus on mechanisms to detect and suppress "Semantic Hallucination" when embedding scores are low.

\section{Conclusion}

This paper presented \textbf{SMoG (Schema Matching on Graph)}, a novel framework that re-establishes the viability of query-based knowledge graph exploration for schema matching. Addressing the limitations of prior vector-based approaches—specifically their "black-box" nature and high storage requirements—we proposed an iterative 1-hop SPARQL exploration strategy.

Our key conclusions are:
\begin{enumerate}
    \item \textbf{Feasibility of Query-based SM:} We proved that query-based methods, when designed as iterative 1-hop explorations, are not only feasible but competitive with SOTA vector-based approaches, debunking previous claims of their inefficiency.
    \item \textbf{Explainability and Efficiency:} SMoG offers a transparent, human-verifiable reasoning path without the overhead of massive vector indices, directly addressing the need for explainable AI in high-stakes medical data integration.
    \item \textbf{Insight into Reasoning Dynamics:} We identified "Over-thinking" and "Premature Commitment" as key failure modes, providing a roadmap for future research to focus on adaptive reasoning control.
\end{enumerate}

By successfully bridging the gap between the transparency of symbolic reasoning and the power of LLMs, SMoG paves the way for more reliable, explainable, and efficient automated data standardization systems. Future work will focus on adaptive depth control to mitigate the specific error types identified, further enhancing the robustness of this approach.

\section*{Ethical Statement}
We use synthetic or de-identified datasets and follow license terms of clinical terminologies. No patient-identifying data is used.

\bibliography{references}
\end{document}